%% file: main.tex
\documentclass[twoside]{article}
\usepackage[accepted]{aistats2022}
\usepackage{graphicx}
\usepackage{bm}
\usepackage{amssymb}
\usepackage[dvipsnames]{xcolor}

\usepackage[round]{natbib}
\renewcommand{\bibname}{References}
\renewcommand{\bibsection}{\subsubsection*{\bibname}}

\usepackage{tikz}
\usetikzlibrary{arrows, decorations.markings, matrix}
\usetikzlibrary{positioning,decorations.pathreplacing,calc}
\usetikzlibrary{bayesnet}
\usepackage{multirow}
\usepackage{standalone}

\usepackage{booktabs}

\usepackage{caption}
\usepackage{subcaption}
\usepackage{hyperref}

\newcommand{\KL}[2]{D_{KL}(#1\,||\,#2)}
\newcommand{\MMD}{\text{MMD}}
\newcommand{\E}{\mathbb{E}}
\newcommand{\z}{\bm{z}}
\renewcommand{\b}{\bm{b}}
\newcommand{\s}{\bm{s}}
\newcommand{\x}{\bm{x}}
\newcommand{\Lmmd}{\mathcal{L}_{\MMD}}
\newcommand{\comment}[1]{}
\newcommand{\vecref}{\bm{s}'}

\definecolor{mypink1}{rgb}{1, 0.4, 0.8}
\definecolor{mybrown1}{rgb}{0.6, 0.2, 0}

\begin{document}

%

%

\twocolumn[

\aistatstitle{Moment Matching Deep Contrastive Latent Variable Models}

\aistatsauthor{Ethan Weinberger \And Nicasia Beebe-Wang \And  Su-In Lee}

\aistatsaddress{University of Washington \And  University of Washington \And University of Washington} ]

\begin{abstract}
In the contrastive analysis (CA) setting, machine learning practitioners are specifically interested in discovering patterns that are enriched in a \textit{target} dataset as compared to a \textit{background} dataset generated from sources of variation irrelevant to the task at hand. For example, a biomedical data analyst may seek to understand variations in genomic data only present among patients with a given disease as opposed to those also present in healthy control subjects. Such scenarios have motivated the development of contrastive latent variable models to isolate variations unique to these target datasets from those shared across the target and background datasets, with current state of the art models based on the variational autoencoder (VAE) framework. However, previously proposed models do not explicitly enforce the constraints on latent variables underlying CA, potentially leading to the undesirable leakage of information between the two sets of latent variables. Here we propose the moment matching contrastive VAE (MM-cVAE), a reformulation of the VAE for CA that uses the maximum mean discrepancy to explicitly enforce two crucial latent variable constraints underlying CA. On three challenging CA tasks we find that our method outperforms the previous state-of-the-art both qualitatively and on a set of quantitative metrics.
\end{abstract}

\section{INTRODUCTION}

In unsupervised learning settings, the goal is to find interesting patterns in a dataset. Here the definition of ``interesting'' is subjective and dependent on the specifics of a given problem domain. In many domains, data analysts are specifically interested in variations that are enriched in one dataset, referred to as the \textit{target}, as compared to a second related dataset, referred to as the \textit{background}. Target and background dataset pairs arise naturally in many settings. For example, data from healthy controls versus a diseased population, from pre-intervention and post-intervention groups, or signal-free versus signal-containing recordings all form natural background and target pairs \citep{abid2018exploring}. In each of these scenarios, the target dataset likely shares some set of uninteresting nuisance variations with the background dataset, such as population level variations or sensor noise. Unfortunately, when these nuisance variations explain the majority of the overall variance in a target dataset, standard unsupervised learning algorithms struggle to recover the patterns of interest in the data.

Isolating these salient variations present only in a target dataset is the subject of \emph{contrastive analysis} (CA; \citep{zou2013contrastive}), and a number of algorithms have been proposed that extend unsupervised learning methods to the CA setting. For example, \citet{abid2018exploring} developed contrastive principal components analysis (cPCA), which finds a set of contrastive principal components that maximize variance in the target dataset while minimizing variance in the background. Moreover, a number of recent works \citep{li2020probabilistic, severson2019unsupervised, abid2019contrastive, ruiz2019learning, jones2021contrastive} have focused on developing methods for learning contrastive latent variable models. In such models, data points are generated using two sets of latent variables: one set, called the \textit{background variables}, is used to generate both target and background data points, while the other, called the \textit{salient variables}, is used exclusively for generating target data points.

In particular, to recover nonlinear relationships between latent variables and observed samples, a number of algorithms have been proposed that adapt the variational autoencoder (VAE) for CA and isolate salient factors of variation in target datasets \citep{abid2019contrastive, severson2019unsupervised, ruiz2019learning}. However, none of these previous methods explicitly enforce the constraints on latent variables assumed by CA during training, thereby possibly learning latent representations that are ineffective for CA tasks.

In this work we explore the impact of enforcing these constraints during model training. The specific contributions of this work are the following:

\begin{enumerate}
    \item We identify two constraints underlying the CA framework that are ignored by previously proposed deep latent variable models for CA.
    \item We propose the moment matching contrastive variational autoencoder (MM-cVAE), a new deep latent variable model for CA that is trained to learn  representations that satisfy these assumptions.
    \item We demonstrate on three challenging datasets that MM-cVAE's representations more closely adhere to the latent variable constraints of CA; moreover, we find that adhering to these constraints results in representations that perform better on downstream CA tasks.
\end{enumerate}

\section{CONTRASTIVE LATENT VARIABLE MODELING}
\label{section:contrastive}

\begin{figure}
    \centering
    \input{target_net_new}
    \caption{Generative processes for target and background samples. Background latent variables $\bm{z}$ are common to the two processes. A set of salient latent variables $\bm{s}$ are also used in the target generative process. For the background dataset these salient variables are fixed at $\vecref$. Observed values are shaded, and square nodes denote constant values.} 
    \label{fig:pgm}
\end{figure}
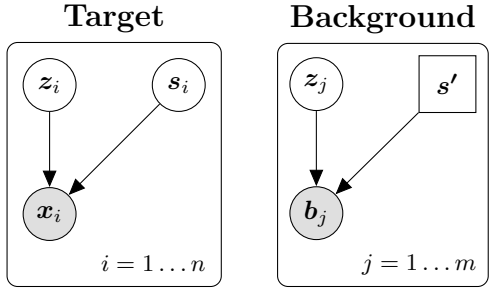

Recall that the goal of CA is to isolate variations only present in a target dataset from those present in both the target and background datasets. To capture this idea, we make use of a contrastive latent variable model defined as follows: let $X = \{\bm{x}_{i}\}_{i=1}^{n}$ denote our target dataset consisting of samples of interest. We assume that our $\bm{x}_i$ are drawn i.i.d.\ from a random process that depends on two sets of latent variables $\bm{z}_i$ and $\bm{s}_i$ distributed according to known prior distributions. Our observed $\bm{x}_i$ are then drawn from a conditional distribution $p_{\theta}$ parameterized by unknown parameters $\theta$:

\begin{align*}
    \bm{x}_i \sim p_{\theta}(\bm{x}|\z_i, \s_i).
\end{align*}

Here $\bm{s}$ refers to a set of \textit{salient} latent variables of interest,
while $\bm{z}$ refers to a set of \textit{background} latent variables that are irrelevant to a given analysis. Our goal in the CA setting is to identify and isolate the salient latent variables for further analysis. Unfortunately, when trained on a target dataset alone, standard unsupervised learning algorithms will not necessarily isolate these salient latent variables. For example, standard VAEs are prone to learning latent representations in which salient latent factors are strongly entangled with background factors \citep{abid2019contrastive, ruiz2019learning, severson2019unsupervised}. Moreover, if the background latent factors explain most of the overall variance in the target dataset, a VAE may fail to capture the salient factors entirely.

Following previous work in the CA literature, we can attempt to solve our problem by introducing a background dataset $B = \{\bm{b}_{j}\}_{j=1}^{m}$ for which we assume that samples are generated only from the irrelevant latent factors with the salient variables fixed to some constant ``reference vector'' $\vecref$ representing a lack of information (e.g., $\bm{0}$). We note that both our target and background samples are assumed to be drawn from the same generative process $p_{\theta}(\cdot)$. Moreover, we do not assume that we have the same number of samples in the target and background datasets (i.e., $n \neq m$), and we do not assume that there is any special relationship between samples $i$ and $j$ from the target and background datasets, respectively. We depict our data generating process in Figure \ref{fig:pgm}. Target and background dataset pairs arise naturally in many experimental settings. For example, suppose we wanted to analyze data from patients receiving an experimental treatment in a clinical trial to better understand differences in how the treatment impacts the patients who receive it. In this case, a suitable background dataset would consist of data from control patients drawn from a similar population but administered a placebo instead of a real treatment, and our salient latent factors of interest would correspond to variations present \textit{only} in patients that received the real treatment.

Our problem is thus to learn the parameters $\theta$ used to generate samples as well as another set of parameters $\phi$ for performing inference of the latent features. In particular, we seek to learn two sets of parameters $\phi_{\s}$ and $\phi_{\z}$ to infer salient and background latent variable values, respectively.

\section{RELATED WORK}
\label{section:related_work}

Since the introduction of the VAE \citep{kingma2014autoencoding}, many works have focused on constraining the representations learned by VAEs to capture factors of variation that are more meaningful to the user. For example, learning disentangled representations \citep{bengio2013representation}, in which each feature encodes a semantically meaningful attribute and is invariant to the other features, is the subject of much previous work \citep{beta-vae, kim2018disentangling, chen2018isolating, mathieu2019disentangling, lopez2018information, kumar2018variational, burgess2018understanding}. Other works have focused on developing semi-supervised variants of the VAE, in which latent factors are encouraged to encode specific, pre-labeled factors of interest \citep{kingma2014semi, paige2017learning}. In contrast, in this work we specifically focus on learning representations that disentangle salient latent variables (which only underlie target data points) and background latent variables (which underlie both target and background data points).

Uncovering such salient variations is the focus of contrastive analysis, originally introduced by \citet{zou2013contrastive}. The CA framework has previously been used to extend a number of unsupervised learning methods. In the initial formulation, \citet{zou2013contrastive} adapted mixture models for CA. Previous work \citep{abid2018exploring} adapted principal component analysis to the contrastive setting so that a set of salient principal components are learned that summarize variations of interest in a target dataset while ignoring uninteresting variations. Other recent works \citep{li2020probabilistic, abid2019contrastive, severson2019unsupervised, ruiz2019learning, jones2021contrastive} have since developed probabilistic latent variable models for CA.

In particular, recent work has focused on adapting VAEs to the CA setting to handle situations in which the relationships between observed data points and latent variable values are highly nonlinear. \citet{severson2019unsupervised} were the first to do so; however, their formulation assumes that target data points can be decomposed as the sum of nonlinear transformations of target and salient latent variable values, i.e., $\x_i = f_{\theta}(\s_i) + g_{\theta}(\z_i)$, where $\s_i$ and $\z_i$ are salient and background latent variable values, respectively, for $\x_i$ and $f_{\theta}$ and $g_{\theta}$ are nonlinear transformations. Such an assumption is highly restrictive, and likely to be violated by complex real-world datasets. Subsequently, two less restrictive models, the cVAE \citep{abid2019contrastive} and sRB-VAE \citep{ruiz2019learning}, were proposed for CA tasks.

\textbf{Shortcomings of previous methods:} Despite their advantages over the original VAE for CA proposed by \citet{severson2019unsupervised}, cVAE and sRB-VAE still have notable shortcomings; namely, their formulations do not fully reflect the constraints on latent variables underlying CA. Specifically, neither model explicitly \textbf{(1)} encourages the distribution of background latent factors to be similar across target and background samples or \textbf{(2)} enforces that inferred salient latent variable values for background samples are close to the informationless reference vector $\vecref$. Such choices may enable information to leak between the salient and background latent spaces (i.e., target-dataset-specific variations may be present in the shared latent space and vice-versa). Such leakage violates the assumptions of the CA framework and may lead to lower quality representations for downstream CA tasks. Moreover, rather than explicitly enforcing these assumptions, cVAE and sRB-VAE add new KL divergence terms to the original VAE objective to encourage disentanglement of salient and background latent factors. Such divergence terms cannot be calculated directly, and they are instead approximated using the density-ratio trick \citep{kim2018disentangling}. However, doing so requires using adversary networks during model training, increasing training time and potentially leading to instabilities during training. In contrast, our MM-cVAE model explicitly enforces the assumptions of contrastive latent variable models during training by using a non-parametric kernel-based measure of independence. Moreover, we find that doing so results in high quality representations without needing the extra KL divergence loss terms used by previous models and the adversary networks needed to approximate them.

\section{METHOD}
\label{section:method}

\begin{figure*}[t]
    \centering
    \includegraphics[width=0.75\linewidth]{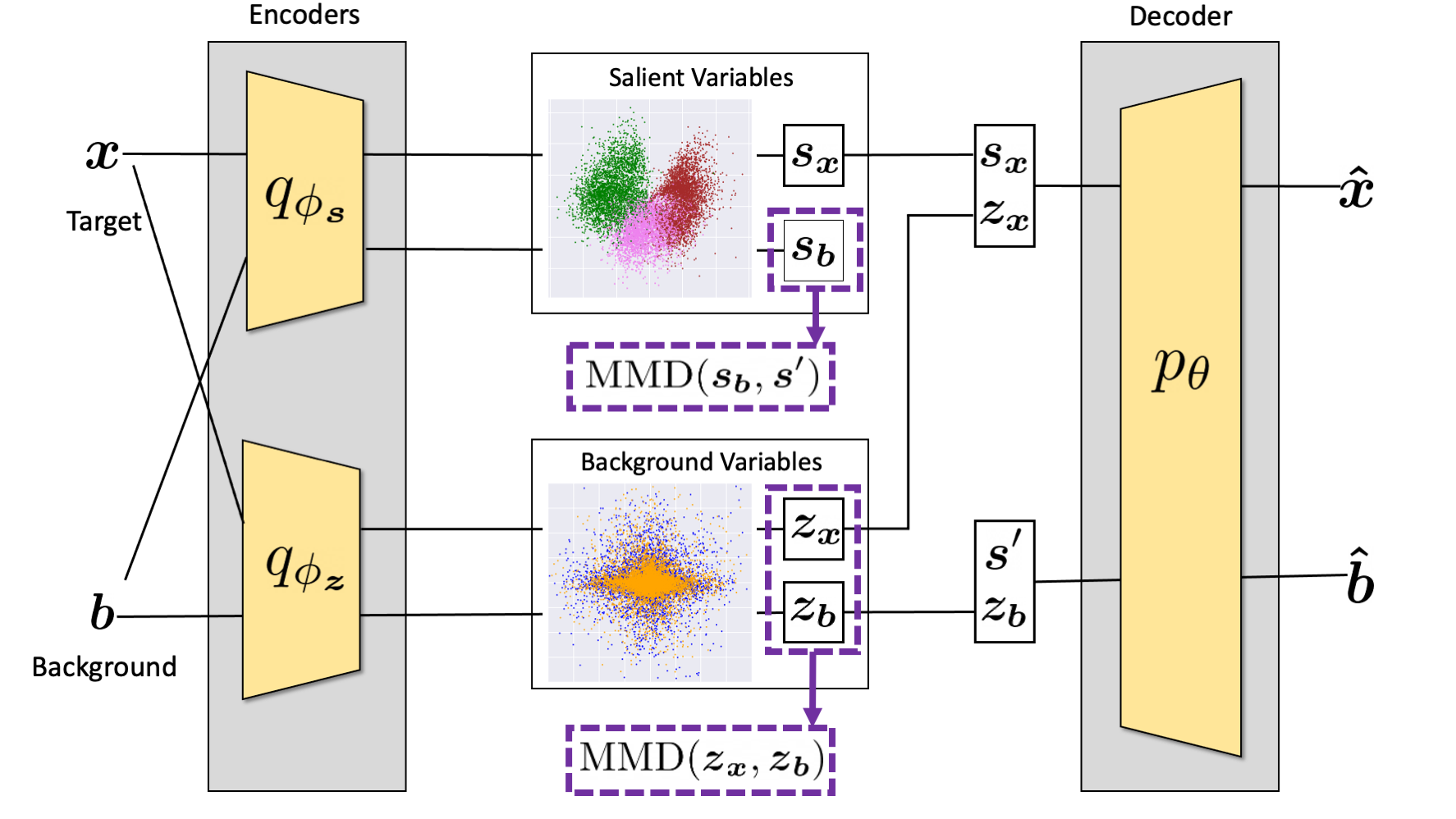}
    \caption{Network architecture for MM-cVAE. Target and background samples are fed through encoder networks $q_{\phi_{\bm{s}}}$ and $q_{\phi_{\bm{z}}}$ to obtain values for the salient and background latent variables, respectively. To explicitly enforce the assumptions of contrastive latent variable modeling, we penalize the Maximum Mean Discrepancy (MMD) between background latent features for background and target samples as well as the MMD between salient features for background samples and a delta distribution centered at an informationless reference vector $\vecref$. \textcolor{blue}{Blue} and \textcolor{orange}{orange} indicate whether background latent variable values are from a target or background sample, respectively. \textcolor{ForestGreen}{Green}, \textcolor{mypink1}{pink}, and \textcolor{mybrown1}{brown} indicate different classes of target samples.}
    \label{fig:concept}
\end{figure*}

Here we present the moment matching contrastive VAE (MM-cVAE, Figure \ref{fig:concept}), our extension of the VAE to the latent variable model introduced in Section \ref{section:contrastive}. Similar to the cVAE \citep{abid2019contrastive}, we use two probabilistic encoders $q_{\phi_{\z}}(\z|\x)$ and $q_{\phi_{\s}}(\s|\x)$ to approximate the posteriors over our two sets of latent variables $\z$ and $\s$, respectively, and we use a single generator network $p_{\theta}(\cdot)$ that takes in a concatenation of $\z$ and $\s$ to reconstruct our data. For target samples, we concatenate the salient and background latent variables $\z_x$ and $\s_x$ as-is when reconstructing data points. However, for background samples, we concatenate $\z_b$ with a constant $\vecref$ to reflect the assumptions of our latent variable model. In our experiments we chose $\vecref = \bm{0}$.

We can then obtain variational lower bounds for the likelihoods of individual data points. For a given point $\x_i$ in our target dataset we have the following bound:\footnote{Originally derived in \citet{abid2019contrastive}. For completeness, we include a full derivation of this bound in the supplement.}

\begin{align*}
\mathcal{L}_{\x}(\x_i) =\ &\mathbb{E}_{q_{\phi_{\z}}(\z)q_{\phi_{\s}}(\s)}[\log p_{\theta}(\x_i | \z, \s)] \\&-
\KL{q_{\phi_{\z}}(\z | \x_i)}{p_{\x}(\z)}\\&- \KL{q_{\phi_{\s}}(\s | \x_i)}{p_{\x}(\s)},
\end{align*}

where $p_{\x}(\z)$ and $p_{\x}(\s)$ are the prior distributions for our background and salient variables respectively for target samples. We let these two prior distributions be multivariate isotropic Gaussians, i.e., $p_{\x}(\z) = \mathcal{N}(\z; \bm{0}, I)$ and  $p_{\x}(\s) = \mathcal{N}(\s; \bm{0}, I)$. For points $\b_j$ in our background dataset we similarly have

\begin{align*}
\mathcal{L}_{\b}(\b_j) =\ &\mathbb{E}_{q_{\phi_{\z}}(\z)}[\log p_{\theta}(\b_j | \z, \vecref)]\\ &- \KL{q_{\phi_{\z}}(\z | \b_j)}{p_{\b}(\z)} \\&- \KL{q_{\phi_{\s}}(\s | \b_j)}{p_{\b}(\s)},
\end{align*}

where $p_{\b}(\z)$ and $p_{\b}(\s)$ are the prior distributions for our background and salient variables respectively for background samples. As with $p_{\x}(\z)$, we let $p_{\b}(\z)$ be an isotropic Gaussian. However, we set $p_{\b}(\s) = \delta\{\s=\vecref\}$, where $\delta\{\s=\vecref\}$ denotes a Dirac distribution centered at $\vecref$, to reflect our assumption that salient latent variables are not used to generate background data points. Unfortunately, this prior distribution makes the second KL divergence term in our expression for $\mathcal{L}_{\b}$ non-differentiable. Prior work \citep{abid2019contrastive, ruiz2019learning} has sidestepped this differentiability issue by ignoring this term altogether during optimization. However, doing so introduces a path for information to leak between the background and salient latent spaces. Rather than ignoring this term altogether, we can instead introduce a differentiable surrogate loss term that encourages the inferred salient latent variable values for background points to be close to $\vecref$.

Here we choose to use the maximum mean discrepancy (MMD)  \citep{gretton2006kernel} as our surrogate. The MMD is a nonparametric kernel-based two sample test statistic originally proposed for testing whether two datasets $X \in \mathbb{R}^{n \times d}$ and $Y \in \mathbb{R}^{m \times d}$ were drawn from the same distribution. For our purposes it is sufficient to know that the MMD has a differentiable closed-form empirical estimator

\begin{align*}
    \label{eq:mmd}
    \Lmmd&(X, Y) = 
    \frac{1}{n^2}\sum_{i=1}^{n}\sum_{j=1}^{n}k(\x_i, \x_j) \\ &-\frac{1}{nm}\sum_{i=1}^{n}\sum_{j=1}^{m}k(\x_i, \bm{y}_j)
    + \frac{1}{m^2}\sum_{i=1}^{m}\sum_{j=1}^{m}k(\bm{y}_i, \bm{y}_j).
\end{align*}

\begin{table*}
    \caption{Quantitative evaluation comparing how well MM-cVAE and two baseline models cVAE \citep{abid2019contrastive} and sRB-VAE \citep{ruiz2019learning} satisfy the assumptions of contrastive latent variable models; i.e., that the distributions of background latent variables $\z_{\x}$ and $\z_{\b}$ should be identical and salient variable values for background samples $\s_{\b}$ should be close to a model's informationless reference vector. To measure distributional similarity, we train logistic regression models to classify points based on their origins and report accuracy on a held out test set. We also report silhouette scores to quantify how well clustered points are based on their origins. Lower values correspond to better satisfying the assumptions.}
    \centering
    \input{table_background}

    \label{table:results_background}
\end{table*}

If $k$ is a universal\footnote{A kernel function $k(x, \cdot)$ is called universal if $k$ is continuous for all $x$ and the RKHS induced by $k$ is dense in $\mathcal{C}(\mathcal{X})$, where $\mathcal{C}(\mathcal{X})$ denotes the space of continuous bounded functions on the compact domain $\mathcal{X}$.} kernel function, such as the Gaussian kernel $k(\x, \bm{y}) = \exp(-\gamma||\x - \bm{y}||^2)$, then minimizing the MMD is equivalent to minimizing a distance between all moments of $X$ and $Y$ \citep{li2015generative}, and the MMD equals 0 asymptotically if and only if $X$ and $Y$ were drawn from the same distribution \citep{gretton2006kernel, gretton2012kernel}. MMD-based penalties using the Gaussian kernel have been successfully used by a number of works to enforce invariance constraints when training deep generative models \citep{shaham2017removal, louizos2016fair, lopez2018information, li2015generative, dziugaite2015training, zhao2019infovae}, and we refer the reader to the supplement for a more detailed treatment of the MMD. Substituting an empirical estimate of the MMD for the undifferentiable divergence term in our equation for $\mathcal{L}_{\b}$ and letting $\mathcal{L}_{\b}'(B)$ denote the new likelihood bound for our entire background dataset $B$ we have,

\begin{align*}
\mathcal{L}_{\b}'(B) = &\sum_{j=1}^{m}\bigg(\mathbb{E}_{q_{\phi_{\z}}(\z)}[\log p_{\theta}(\b_j | \z, \vecref)] \\ &-\KL{q_{\phi_{\z}}(\z | \b_j)}{p_{\b}(\z)}\bigg) \\ &-\lambda_{1}\cdot\Lmmd(\hat{q}_{\phi_{\s, \b}}(\s), \delta\{\s=\vecref\}),
\end{align*}

where $\hat{q}_{\phi_{\s, \b}}(\s)$ denotes the empirical distribution of inferred salient variable values for background data points and $\lambda_{1}$ is a hyperparameter controlling regularization strength. 

An additional assumption of our contrastive latent variable model is that the distribution of common latent variables $\z$ is identical across our background and target datasets. Previous work \citep{abid2019contrastive, ruiz2019learning} ignores this constraint, adding yet another potential avenue for information to leak between the salient and background latent spaces. Here we explicitly enforce this assumption by once again making use of the MMD. Let $\hat{q}_{\phi_{\z, \x}}(\z)$ and $\hat{q}_{\phi_{\z, \b}}(\z)$ denote the empirical distributions of shared latent factors for our target and background datasets, respectively. We then add a penalty term $-\lambda_{2} \cdot \Lmmd(\hat{q}_{\phi_{\z, \x}}(\z),\,\hat{q}_{\phi_{\z, \b}}(\z))$ to our objective function, where $\lambda_2$ is another hyperparameter controlling regularization strength. Letting $\mathcal{L}_{\x}(X) = \sum_{i=1}^{n}\mathcal{L}_{\x}(\x_i)$, we then have our final objective

\begin{align*}
    \max_{\theta, \phi}  \mathcal{L}_{\x}(X) + \mathcal{L}_{\b}'(B)
    &-\lambda_{2} \cdot \Lmmd(\hat{q}_{\phi_{\z, \x}}(\z),\,\hat{q}_{\phi_{\z, \b}}(\z)),
\end{align*}

which we can optimize using a variant of minibatch gradient descent. Due to the use of the MMD in our objective function, we call our approach the moment matching contrastive VAE (MM-cVAE).

\section{EXPERIMENTS}
\label{section:results}

To confirm that MM-cVAE's loss function results in higher-quality representations for CA, we experimented with three collections of target and background datasets (Section \ref{subsec:datasets}). For each of these collections, points in the target dataset have ground truth class labels. However, distinguishing between these classes may be difficult from latent variable values inferred using a non-contrastive approach (e.g.\ a standard VAE). In our experiments, we compared our MM-cVAE with cVAE \citep{abid2019contrastive} and sRB-VAE \citep{ruiz2019learning}, the previous state of the art deep latent variable models for CA. For each dataset, we evaluated our models' performance quantitatively and qualitatively (Section \ref{subsection:results}). We refer the reader to the supplement for additional details on dataset preprocessing, model training, and hyperparameter tuning. The code used to preprocess our datasets as well as to train and evaluate our models is available at \url{https://github.com/suinleelab/MM-cVAE}.
 
\subsection{Datasets}
\label{subsec:datasets}

\textbf{CelebA images with accesories.} We created target and background datasets from subsets of the well-known CelebA \citep{liu2015deep} dataset. Our target dataset comprised images of celebrities wearing either glasses or caps, while our background dataset contained images of celebrities wearing neither accessory. Our goal here was to learn a salient representation of the target dataset that separated celebrities wearing glasses versus those with caps.  

\noindent
\textbf{Leukemia treatment response.} Here we combined datasets from \citet{zheng2017massively} containing gene expression measurements from two patients with acute myeloid leukemia (AML) before and receiving a blood cell transplant. Our goal here is to learn a representation that separates pre- and post-transplant measurements. As a background dataset we used expression measurements from two healthy control patients that were collected as part of the same study.

\noindent
\textbf{Epithelial cell infection response.} We constructed our target dataset by combining two sets of gene expression measurements from \citet{haber2017single}. These datasets consist of gene expression measurements of intestinal epithelial cells from mice infected with either \textit{Salmonella} or \textit{Heligmosomoides polygyrus (H. poly)}. Here our goal is to separate cells by infection type. As a background dataset we used measurements collected from healthy cells released by the same authors. 

\subsection{Assessing Adherence to the Assumptions of CA}
\label{subsection:results}

\begin{figure}
     \centering
     \begin{subfigure}[b]{0.49\linewidth}
         \centering
         \includegraphics[width=\textwidth]{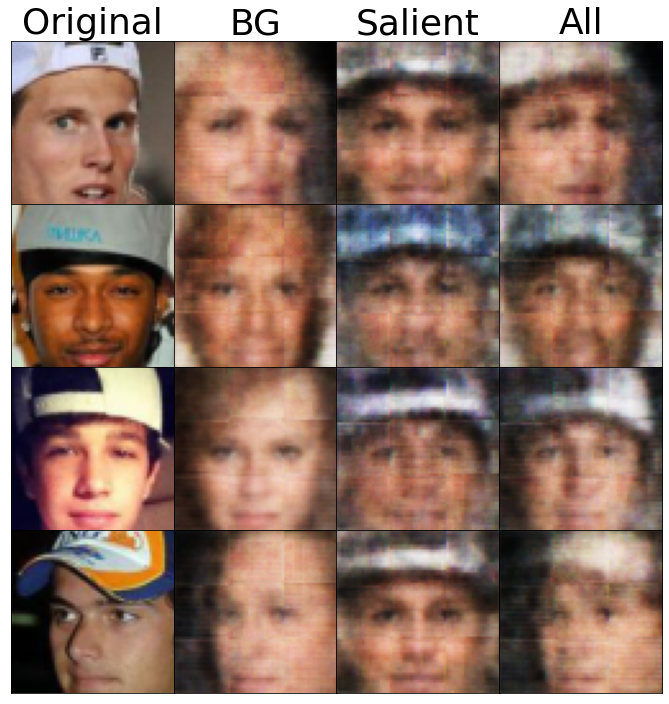}
         \caption{}
         \label{fig:y equals x}
     \end{subfigure}
     \begin{subfigure}[b]{0.49\linewidth}
         \centering
         \includegraphics[width=\textwidth]{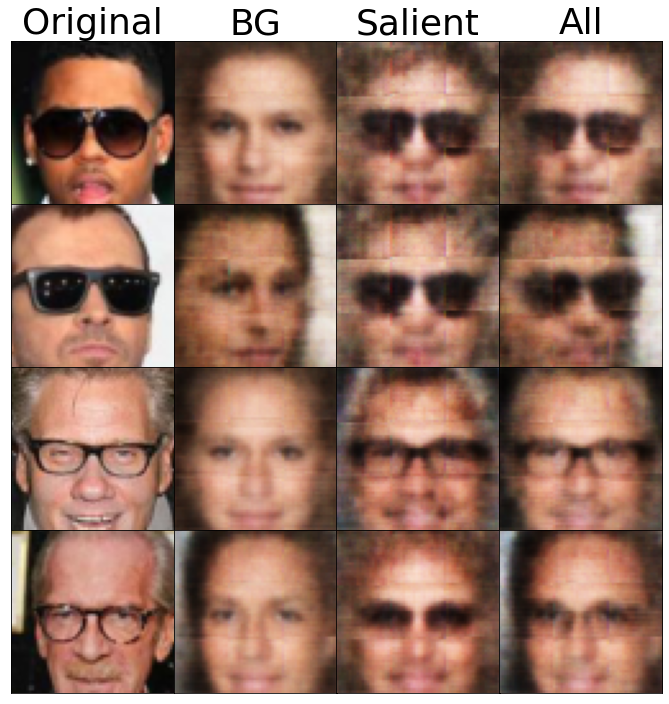}
         \caption{}
         \label{fig:three sin x}
     \end{subfigure}

        \caption{Sample of qualitative results on the CelebA with accessories task for images with hats (\textbf{a}) and glasses (\textbf{b}). For both sets of samples we present original images (left), reconstructions using only background (BG) variables (center left), reconstructions using only salient variables (center right), and reconstructions using all latent variables (right). For both collections of images, the salient latent variables encode the accessory of interest (i.e., hats or glasses) along with other seemingly random facial features. On the other hand, the background variables encode the remaining attributes (e.g., pose, lighting etc.)}
        \label{fig:celeba_qualitative}
\end{figure}

\begin{figure*}[!ht]
    \centering
    \includegraphics[width=\linewidth]{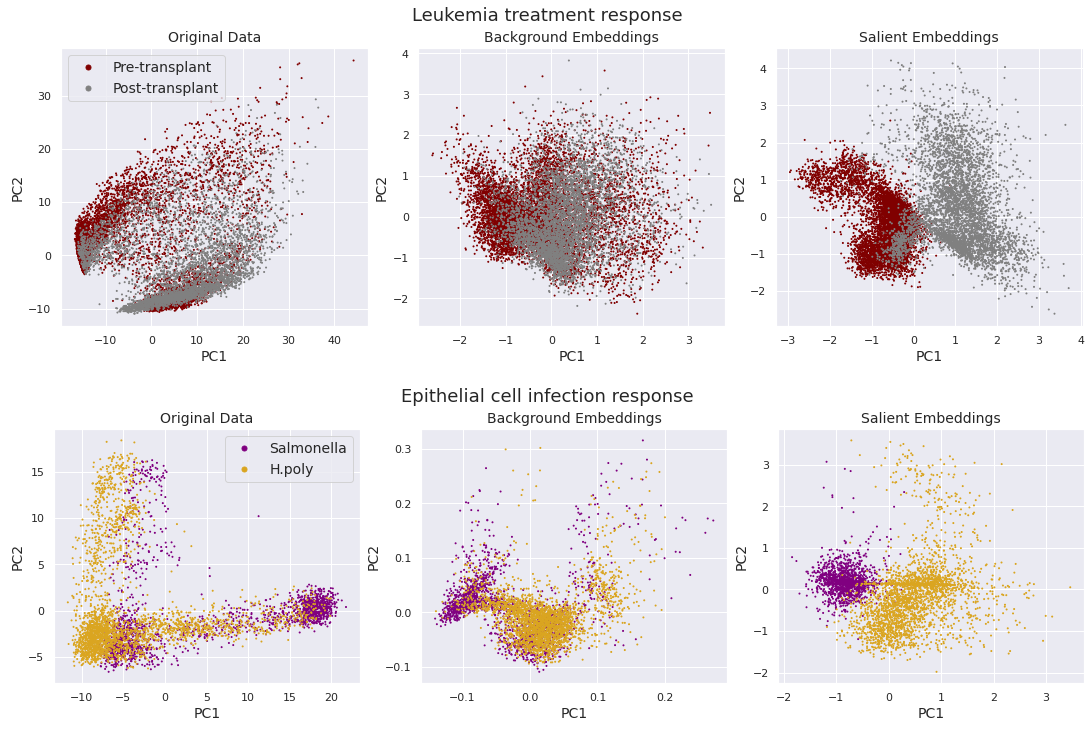}
    \caption{Plots of the first two principal components for our target RNA-seq datasets pre-embedding (left), embedded in the shared latent space (center) and in the salient embedding space (right). For both target datasets, the two subgroups are difficult to distinguish in the spaces spanned by the first two principal components for the original data and the background embeddings. However, the subgroups are more easily separable in the salient embedding space.}
    \label{fig:qualitative_rna}
\end{figure*}

We began our evaluation by quantitatively assessing how well MM-cVAE and our baseline models' learned representations satisfy the assumptions of contrastive latent variable modeling. First, we evaluated how well each model satisfies the assumption that the shared background latent variables for background and target samples have the same distribution (i.e., the assumption that $\z_{\b}$ and $\z_{\x}$ have the same distribution). To do so, we trained logistic regression classifiers on each models' background latent variable values to classify whether a data point came from a background sample or target sample. We also used the Silhouette Score (SS) to quantify how well $\z_{\x}$ and $\z_{\b}$ separated into distinct clusters. In short: the SS varies between -1 and 1, with larger positive values corresponding to better separated clusters and values near 0 indicating overlapping clusters; we refer the reader to the supplement for additional details on the SS. A model that satisfies this assumption, producing similar distributions of $\z_{\x}$ and $\z_{\b}$, should result in low values for both of these metrics. Using a similar methodology we also quantified how well each model satisfied the assumption that inferred salient latent variable values for background samples $\s_{\b}$ should be similar to models' reference vectors $\vecref$. That is, we trained logistic regression classifiers to distinguish between models' values of $\s_{\b}$ and copies of $\vecref$, and we used the SS to quantify how well each model's values of $\s_{\b}$ and copies of $\vecref$ formed distinct clusters. Once again, lower values correspond to better satisfying the assumption. 

We report the values of these metrics for both assumptions in Table \ref{table:results_background}. We find that MM-cVAE consistently outperforms our baseline models on these metrics, often considerably. These results indicate that MM-cVAE's loss function leads it to learn representations that align more closely with the assumptions of contrastive latent variable modeling than do those of previously proposed models. Given these results, we next confirmed that learning representations that better conform to the assumptions of contrastive latent variable modeling indeed leads to better performance on downstream CA tasks.

\subsection{Assessing Latent Representation Quality}
 We next assessed the quality of each model's representations for downstream CA tasks. We began our assessment by qualitatively evaluating these representations. For the CelebA dataset, we did so as follows. For a target image $\x$, we first inferred $\x$'s background latent variable values by feeding it into $q_{\phi_{\b}}$. Next, we concatenated these inferred background features with a zero vector in place of the salient variable values, and we then passed these values through the decoder network to generate a new image. We also performed the same procedure using salient variable values inferred by $q_{\phi_{\s}}$ and a zero vector in place of the background variable values. If a model has learned representations that are well-suited for CA, we would expect it to generate faces with the accessory present in the original target image (i.e., glasses or a hat) and arbitrary settings for other attributes when decoding using only salient variable values; similarly, we would expect the model to generate faces with similar characteristics (e.g., pose, lighting) as the original target image but with no accessory present when decoding using only background variables. Since a neural network may transform an input of $\bm{0}$ to a non-zero output, in these experiments we constrained our decoder network's bias terms to all be zero. We report results from this procedure in Figure \ref{fig:celeba_qualitative}.

\begin{table*}
    \caption{Quantitative comparison of MM-cVAE and two baseline models cVAE \citep{abid2019contrastive} and sRB-VAE \citep{ruiz2019learning}. To measure the quality of a model's representations, we train logistic regression classifiers to predict target data points' classes from their background and salient representations. We also use the silhouette score to measure how well separated points from different classes are. Higher (lower) values on salient (background) representations correspond to better performance.}
    \centering
    \input{table_target}
    \label{table:results_target}
\end{table*}

From visually inspecting the images generated using only background variables, we find that our background latent variables appear to encode sensible values for non-accessory-related attributes while not encoding information on the accessories themselves. Moreover, by inspecting the images generated from salient variables, we find that our salient variables consistently encode the correct accessory while simultaneously encoding arbitrary values for other attributes. On the other hand, images produced with our baseline models using the same procedure (supplement) indicate that these models were less effective at separating salient and target factors of variation, with information appearing to leak between the two latent spaces (e.g., accessories appear in images generated solely from background variables). Taken together, these results indicate that our training procedure does indeed result in representations that are better suited for CA than those of previously proposed models.

We next turned our attention to the two gene expression datasets. For these datasets, we could not easily assess the quality of generated samples by visually inspecting them as we did with CelebA. Thus, we adopted a different procedure. For both datasets we used $q_{\phi_{\b}}$ and $q_{\phi_{\s}}$ to infer values for our background and salient variables for our target data points. We then performed principal component analysis separately on the inferred background and salient latent variable values, and we show plots of the first two principal components for both in Figure \ref{fig:qualitative_rna}. From our plots, we can observe strong mixing between the two known subgroups for both datasets in the background latent variable space. This phenomenon indicates that our background latent variable spaces are encoding shared features across the groups, as desired. On the other hand, we find clear separation between subgroups in the plots of our salient latent variables. We performed the same procedure using our baseline models, and we provide the results in the supplement. Once again, we find that our baselines are less effective at disentangling salient and background latent variables, with stronger (undesirable) separation between subgroups in the background latent variable space and mixing between subgroups in the salient space; these phenomena are especially pronounced for the Epithelial cell dataset. These results further indicate that MM-cVAE learns representations that are better suited for CA.

We next quantitatively evaluated how well each model's salient representations for target points $\s_{\x}$ separated classes within each target dataset (e.g., glasses vs.\ caps in CelebA). To do so, we used the same metrics as in Table \ref{table:results_background}. 
That is, we trained logistic regression classifiers to classify points based on ground-truth class labels and used silhouette scores to measure how well clustered points from different classes were. As we want classes to be well separated in the salient embedding space, \textit{higher} values here correspond to better performance. We also computed these metrics for the common latent factors $\z_{\x}$. Because our background latent factors should have similar distributions across the classes within each target dataset, \textit{lower} values for our metrics on $\z_{\x}$ correspond to better performance. We report these results in Table \ref{table:results_target}. We find that MM-cVAE once again consistently outperforms both baseline models across all our metrics. These results provide further evidence that MM-cVAE's stronger adherence to the assumptions underlying CA indeed leads to improved performance on downstream CA tasks.

\subsection{Assessing Sample Quality}

In addition to latent representation quality, we also assessed the quality of samples generated by each model. In particular, we sought to understand whether MM-cVAE's better disentanglement of salient and background variables came at the expense of sample quality. For the CelebA dataset we computed the Frechet Inception distance (FID; \citep{heusel2017gans}) for each model, and we report our results in Supplementary Table \ref{table:fid}. We computed the FID separately for generated background samples and for generated target samples. Since FID scores for all contrastive models were relatively high - indicating low sample quality across all contrastive methods tested - we also evaluated a non-contrastive VAE with otherwise identical architecture. Its scores were also high, indicating that the high scores were due to our relatively simple architecture (as opposed to e.g.\ VQ-VAE \citep{van2017neural}), rather than the contrastive setup. We also evaluated sample quality for our gene expression datasets using MMD in place of FID (Supplementary Table \ref{table:mmd}). MM-cVAE performed the best for most metrics, suggesting MM-cVAE achieves better representation quality while maintaining similar sample quality to previously proposed models.

\section{DISCUSSION}
In this work we demonstrated that previous VAE-based models designed for CA learn representations that often in fact strongly violate the assumptions of CA. In response to this issue, we proposed MM-cVAE. In contrast to previous work, MM-cVAE explicitly enforces the assumptions of CA by incorporating two penalties based on the maximum mean discrepancy into the training process. Moreover, by doing so, we avoid the need to use complex adversarial training procedures used by previously proposed models to disentangle salient and background latent variables.

Through both qualitative and quantitative evaluations on the CelebA dataset and two real-world biological datasets, we find that explicitly enforcing these penalties led to higher quality embeddings for CA tasks, effectively stratifying different classes of target samples and relegating irrelevant latent factors to a separate background embedding space.

\subsubsection*{Acknowledgements}
We would like to thank Ian Covert, members of the AIMS Lab at the University of Washington, and our AISTATS reviewers for feedback that greatly improved this work. We would also like to thank the Theis Lab at Helmholtz Munich for their development of high-quality open-source software for analyzing single-cell RNA-sequencing data (\texttt{scanpy}), which facilitated this work. This work was funded by the National Science Foundation [DGE-2140004, DBI-1552309, and DBI-1759487] and the National Institutes of Health [R35 GM 128638, and R01 NIA AG 061132].

\bibliography{main}


\clearpage
\appendix

\thispagestyle{empty}

\renewcommand{\figurename}{Supplementary Figure}
\renewcommand{\tablename}{Supplementary Table}
\setcounter{figure}{0}
\setcounter{table}{0}

\onecolumn \makesupplementtitle

\section{Variational Lower Bound Derivations}

Here we present derivations for the likelihood lower bounds presented in the main text. For our target data points, we have

\begin{align*}
    \log p_{\theta}(\x_i) &= \log\int p_{\theta}(\x_i, \s, \z)d\s d\z \\
    &= \log \int \frac{p_{\theta}(\x_i, \s, \z)\cdot q_{\phi}(\x_i | \s, \z)}{q_{\phi}(\x_i | \s, \z)}d\s d\z \\
    &\geq \int q_{\phi}(\x_i | \s, \z) \log \frac{ p_{\theta}(\x_i, \s, \z)}{q_{\phi}(\x_i | \s, \z)}d\s d\z \\
    &= \int q_{\phi}(\x_i | \s, \z) \log \frac{ p_{\theta}(\x_i | \s, \z)\cdot p(\s, \z)}{q_{\phi}(\x_i | \s, \z)}d\s d\z \\
    &= \int \bigg( q_{\phi}(\x_i | \s, \z) \log p_{\theta}(\x_i | \s, \z) + \log \frac{p(\s, \z)}{q_{\phi}(\x_i | \s, \z)}\bigg) d\s d\z \\
    &= \E_{q_{\phi}(\s, \z)}\big[\log p_{\theta}(\x_i | \s, \z)\big] - \KL{q_{\phi}(\s, \z | \x_i)}{p(\s, \z)} \\
    &= \E_{q_{\phi}(\s) q_{\phi}(\z)}\big[\log p_{\theta}(\x_i | \s, \z)\big] - \KL{q_{\phi}(\s | \x_i)}{p(\s)} - \KL{q_{\phi}(\z | \x_i)}{p(\z)} \\
    &\triangleq \mathcal{L}_{\x}(\x_i)
\end{align*}

where we obtain the inequality in line three of our derivation using Jensen's inequality, and our last line follows from the assumed independence of our salient and target latent values $\s$ and $\z$. Here we take our prior distributions $p(\s)$ and $p(\z)$ to be isotropic Gaussians. $\mathcal{L}_{\b}$ follows essentially the same derivation, with $p(\z)$ chosen to be an isotropic Gaussian and $p(\s)$ a delta distribution centered at some reference vector $\bm{s}'$.

\section{Dataset Preprocessing}
\label{appendix:preprocessing}

Here we provide preprocessing details for each of the datasets used in Section \ref{subsec:datasets}. Jupyter notebooks with implementations of our preprocessing workflows can be found in the supplementary materials.\\

\noindent
\textbf{Epithelial Cell}. We began by downloading \texttt{GSE92332\_SalmHelm\_UMIcounts.txt} from \url{https://www.ncbi.nlm.nih.gov/geo/query/acc.cgi?acc=GSE92332}. Each column in this file corresponds to a single cell sample, while each row corresponds to the expression level of a given gene. For each cell, we extracted its condition (i.e., healthy control or disease status) from that cell's label in the first row of the file. In particular, in our experiments we used cells labelled as \texttt{Control},\ \texttt{Salmonella},\ and \texttt{Hpoly.Day10}; cells with other labels were discarded.

Following standard practices in single cell RNA-seq analyses, we first normalized the counts such that each cell had a total count of 10000 after normalization. We then took the logarithm of these normalized counts + 1. For these two normalization steps we used the \texttt{normalize\_total} and \texttt{log1p} functions implemented in the \texttt{scanpy} Python package. Finally, we filtered our data to only have the features selected by the \texttt{highly\_variable\_genes} function in the \texttt{scanpy} Python package.\\

\noindent
\textbf{Leukemia Treatment Response}. We downloaded a set of scRNA-seq gene expression measurements from a leukemia patient pre and post-transplant from \url{https://cf.10xgenomics.com/samples/cell-exp/1.1.0/aml027_pre_transplant/aml027_pre_transplant_filtered_gene_bc_matrices.tar.gz} and \url{https://cf.10xgenomics.com/samples/cell-exp/1.1.0/aml027_post_transplant/aml027_post_transplant_filtered_gene_bc_matrices.tar.gz}, respectively. Similar measurements were downloaded for a second patient at \url{https://cf.10xgenomics.com/samples/cell-exp/1.1.0/aml035_pre_transplant/aml035_pre_transplant_filtered_gene_bc_matrices.tar.gz} and \url{https://cf.10xgenomics.com/samples/cell-exp/1.1.0/aml035_post_transplant/aml035_post_transplant_filtered_gene_bc_matrices.tar.gz}. A set of measurements from a healthy control patient were also downloaded from \url{https://cf.10xgenomics.com/samples/cell-exp/1.1.0/frozen_bmmc_healthy_donor1/frozen_bmmc_healthy_donor1_filtered_gene_bc_matrices.tar.gz}.

We then preprocessed our data in a similar manner as done in previous work \citep{li2020probabilistic}. That is, we took the log of expression counts + 1, removed any gene features or cell samples consisting entirely of zeros, and filtered down the list of gene features for each file to those that were shared among all the files. Finally, we filtered our data down to the 1000 highest variance gene features.\\

\noindent
\textbf{CelebA}. We cropped each image and scaled it to be 64x64 pixels. We then constructed our target dataset as follows. First, we filtered our images to those with a 1 value for the binary \texttt{Eyeglasses} attribute and a 0 value for the binary \texttt{Wearing\_Hat} attribute (i.e., the people in these images are wearing glasses and not hats). We then randomly choose 5000 of these images. Next, we filtered our images to those with a 1 value for the binary \texttt{Wearing\_Hat} attribute and a 0 value for the binary \texttt{Eyeglasses} attribute (i.e., the people in these images are wearing hats and not glasses). We then selected 5000 of these images, and combined them with the previously selected images to form our target dataset. For our background dataset we randomly selected 10000 images with both binary attributes having a 0 value.

\section{Implementation Details and Hyperparameter Tuning}

All models were implemented using PyTorch with the PyTorch Lightning framework.\footnote{https://github.com/PyTorchLightning/pytorch-lightning} All models were optimized using ADAM \citep{kingma2014adam} with a learning rate of 0.001, $\beta_{1} = 0.9$ and $\beta_{2} = 0.999$. Code implementing all models is available as part of the supplementary materials.\\

\noindent
\textbf{Network Architectures:}
For our experiments with the two genomic datasets, each model's encoder networks had a single fully connected hidden layer with 400 units. A ReLU activation function was used at the hidden layer. These hidden layers then connected to salient/background latent spaces with 5 and 10 nodes, respectively. Our decoder networks consisted of the same architecture in reverse. For cVAE, we used separate encoder networks for background and salient latent variables while we used a single network for sRB-VAE as described in the models' original papers. For MM-cVAE we took the same approach as cVAE (i.e., using separate encoder networks). For all models we used a single decoder network to map from the latent variables to the original input space.

For the CelebA experiments, our encoder networks instead consisted of four convolutional blocks. Each block contained a convolutional layer with a 4x4 kernel and 2x2 stride with the padding parameter set to maintain the original input size. This convolutional layer was then followed by a LeakyReLU activation function with a negative slope of 0.2 and a batch normalization layer. The number of output channels in each convolutional layer were 32, 64, 128 and 256, respectively. After the convolutional blocks, the outputs were connected via fully connected layer to the salient/background latent spaces. For each model we used a salient latent space size of 5 and background latent space of size 15. Our decoder networks consisted of the same convolutional blocks architecture in reverse.\\

\noindent
\textbf{Hyperparameter choices for MM-cVAE:} To tune MM-cVAE's first regularization parameter $\lambda_1$ (i.e., the hyperparameter controlling the strength of the KL surrogate term), we set its value so that the resulting loss term would be similar in magnitude to that of the other KL term $\sum_{j}\KL{q_{\phi_{\z}}(\z | \b_j)}{p_{\b}(\z)}$ for background data points in a given minibatch. We found that setting $\lambda_1 = 10e2$ accomplished this goal in all our experiments.

We set $\lambda_2$'s value to keep the scale of our second penalty similar to that of the lower bound, as done with another MMD-based penalty proposed in previous work \citep{louizos2016fair}. We found that setting $\lambda_2 = 10e3$ accomplished this goal for all our experiments.

\begin{figure}
    \centering
    \includegraphics[width=\textwidth]{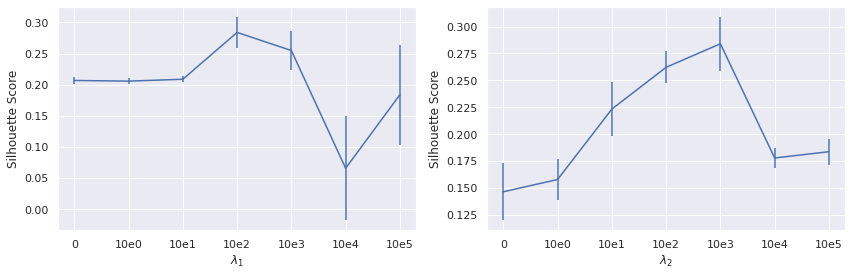}
    \captionof{figure}{Sensitivity analysis on for MM-cVAE's salient latent space for the AML gene expression dataset.}
    \label{fig:sensitivity_AML}
\end{figure}

\begin{figure}
    \centering
    \includegraphics[width=\textwidth]{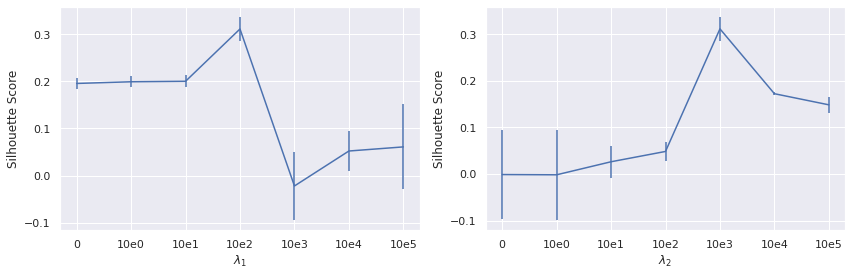}
    \captionof{figure}{Sensitivity analysis on for MM-cVAE's salient latent space for the mice epithelial cell gene expression dataset.}
    \label{fig:sensitivity_epithelial}
\end{figure}

We further explored the impact of these hyperparameters by performing a sensitivity analysis of MM-cVAE's performance with respect to $\lambda_1$ and $\lambda_2$ on salient representations for the two gene expression datasets (Supplementary Figure \ref{fig:sensitivity_AML}, Supplementary Figure \ref{fig:sensitivity_epithelial}). At low values of $\lambda_1$ and $\lambda_2$, our MMD loss terms are dominated by other terms and do not have a strong effect on performance. We found that increases in both parameters up until the values described previously improved representation quality as the MMD loss terms were no longer ignored during training. Once these terms began dominating the loss, representation quality suffered.

\section{The Maximum Mean Discrepancy}

Suppose we have two datasets $X = \{x_{i}\}_{i=1}^{N} \sim P_{0}$ and $Y = \{y_{j}\}_{j=1}^{M} \sim P_{1}$, and we wish to test whether the datasets were drawn from the same distribution; i.e., whether $P_{0} = P_{1}$. One way to do so is via a test statistic known as the maximum mean discrepancy (MMD) \citep{gretton2012kernel}. The MMD is based on the following intuitive idea: if the empirical statistics $\psi(\cdot)$ of the two datasets are similar, then the datasets were likely to have been drawn from the same distribution. Formally, an empirical estimator for the MMD is defined as

\begin{align*}
    \widehat{\MMD}(X, Y) &= \bigg|\bigg|\frac{1}{N}\sum_{i=1}^{N}\psi(x_{i}) - \frac{1}{M}\sum_{j=1}^{M}\psi(y_j) \bigg|\bigg|^2 \\
    &= \frac{1}{N^2}\sum_{i=1}^{N}\sum_{i'=1}^{N}\psi(x_i)^{T}\psi(x_{i'}) - \frac{2}{NM}\sum_{i=1}^{N}\sum_{j=1}^{M}\psi(x_i)^{T}\psi(y_j) + \frac{1}{N^2}\sum_{j=1}^{M}\sum_{j'=1}^{M}\psi(y_j)^{T}\psi(y_{j'})
\end{align*}

Because each term in the previous equation only depends on the inner products of the $\psi$ vectors, we can rewrite it using the kernel trick to obtain

\begin{align*}
    \widehat{\MMD}(X, Y) = \frac{1}{N^2}\sum_{i=1}^{N}\sum_{i'=1}^{N}k(x_i, x_{i'}) - \frac{2}{NM}\sum_{i=1}^{N}\sum_{j=1}^{M}k(x_i, y_j) + \frac{1}{N^2}\sum_{j=1}^{M}\sum_{j'=1}^{M}k(y_{j}, y_{j'})
\end{align*}

 For a universal\footnote{A kernel function $k(x, \cdot)$ is called universal if $k$ is continuous for all $x$ and the RKHS induced by $k$ is dense in $\mathcal{C}(\mathcal{X})$, where $\mathcal{C}(\mathcal{X})$ denotes the space of continuous bounded functions on the compact domain $\mathcal{X}$.} kernel such as the Gaussian kernel $k(x, y) = e^{-\gamma ||x - y||^2}$, we can use a Taylor expansion to obtain an explicit feature map $\psi$ that contains an infinite number of terms and which covers all orders of statistics. Minimizing the MMD for such a kernel function is equivalent to minimizing the distance between all moments of $P_{0}$ and $P_{1}$, and it has been shown \citep{gretton2006kernel} that asymptotically $\widehat{\MMD}(X, Y) = 0$ if and only if $P_{0} = P_{1}$.
 
 \section{Qualitative Results for Baseline Models}

 \subsection{CelebA}
 
 \begin{figure}[h]
     \centering
     \begin{subfigure}[b]{0.4\textwidth}
         \centering
         \includegraphics[width=\textwidth]{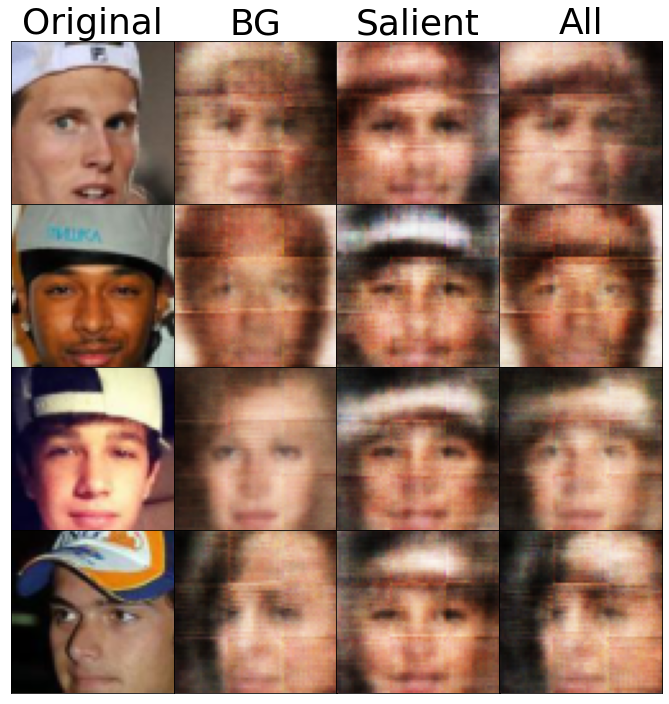}
         \caption{}
     \end{subfigure}
     \begin{subfigure}[b]{0.4\textwidth}
         \centering
         \includegraphics[width=\textwidth]{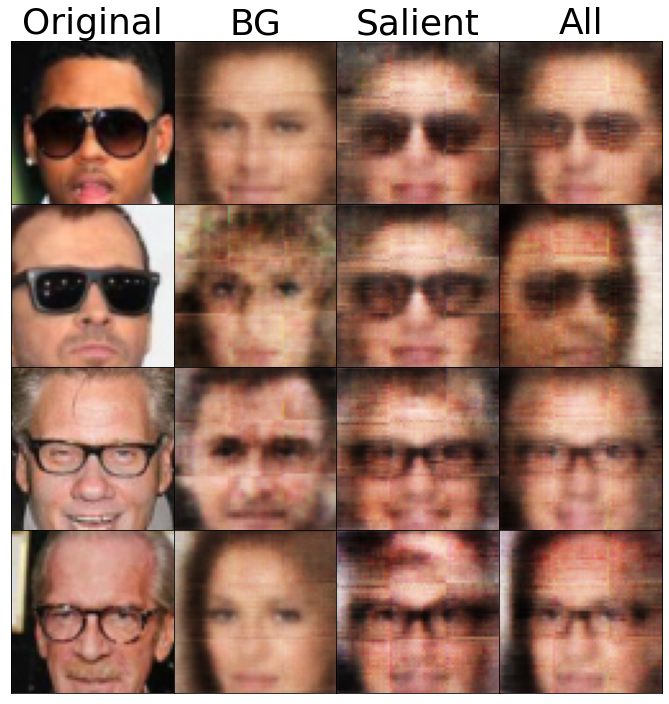}
         \caption{}
     \end{subfigure}
        \caption{Images generated with cVAE using the same procedure as in Figure 3 in the main text. From a visual inspection, we can see that salient and uninteresting factors of variation are not as well disentangled for these iamges as they were with MM-cVAE.}
    \label{fig:celeba_qualitative_cvae}
\end{figure}
 
 We first present qualitative results for the CelebA dataset for cVAE (Supplementary Figure \ref{fig:celeba_qualitative_cvae}), generated using the same procedure described in the main text for MM-cVAE. From these results, it appears that cVAE fails to disentangle salient and background latent factors as effectively as MM-cVAE. For example, a clear outline of a hat is present in the topmost image in the background column in Supplementary Figure 3a. Similarly, dark glasses-like circles are present near the eyes in the BG column for the image in the row second from the bottom of Supplementary Figure 3b.
 
 \begin{figure}[h]
    \centering
    \begin{subfigure}[b]{0.4\textwidth}
        \centering
        \includegraphics[width=\textwidth]{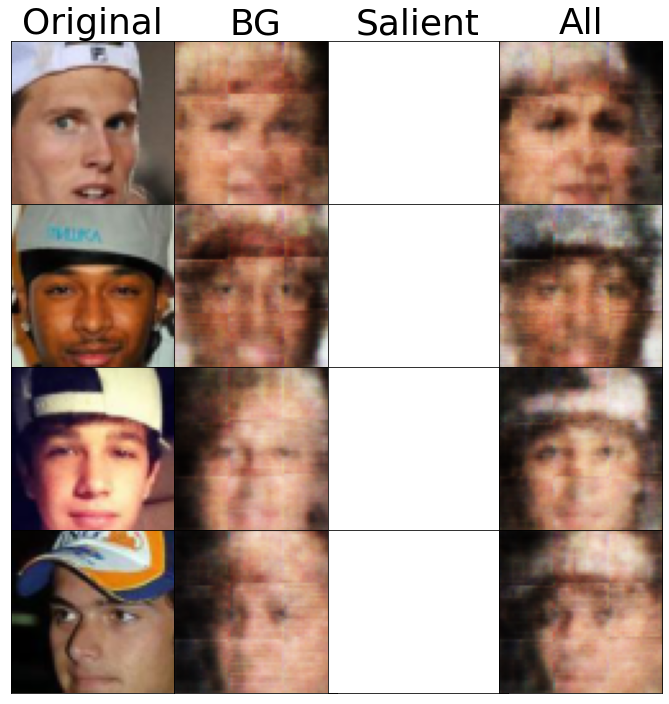}
        \caption{}
    \end{subfigure}
    \begin{subfigure}[b]{0.4\textwidth}
        \centering
        \includegraphics[width=\textwidth]{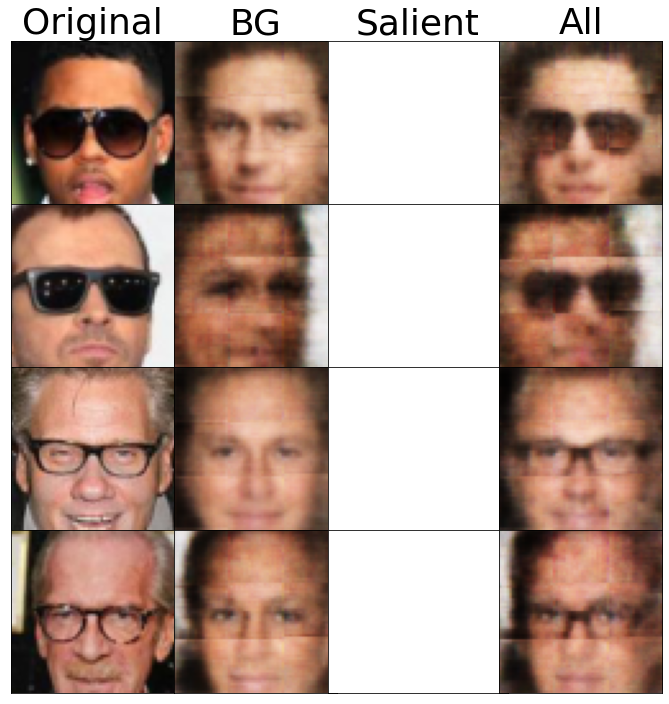}
        \caption{}
    \end{subfigure}
        \caption{Images generated with sRB-VAE using a slightly modified version of the procedure as in Figure 3 in the main text. Once again, we can see that salient and uninteresting factors of variation are not as well disentangled for these images as they were with MM-cVAE.}
    \label{fig:celeba_qualitative_symmetric}
\end{figure}

We next generated images using sRB-VAE (Supplementary Figure \ref{fig:celeba_qualitative_symmetric}). Due to the details of the sRB-VAE architecture, we were forced to make some minor modifications to our image generation procedure. Rather than use a combination of zero vectors and zero bias terms to indicate the absence of features, sRB-VAE uses a learned vector of constants $\bm{s}'$ to represent uninformative salient variable values for background data points. Thus, to generate images using only background variables, we concatenated images' inferred background variable values with this reference vector before decoding. From these generated images, we can see that sRB-VAE too seems to struggle with disentangling salient and background latent factors. For example, images generated using background latent variable values for celebrities with hats (Supplementary Figure \ref{fig:celeba_qualitative_symmetric}a) all appear to have outlines of hats. Similarly, we can see glasses start to form in some images generated using background variables from celebrities with glasses. Specifically, we see sunglasses-like dark circles in our image in the second row of Supplementary Figure \ref{fig:celeba_qualitative_symmetric}b, and faint glasses frames forming in the fourth row image.

Unfortunately, sRB-VAE does not provide a corresponding vector of uninformative values for background latent variables, preventing us from generating samples using only salient latent variable values as we did with cVAE and MM-cVAE.

\subsection{RNA-Seq data}

\begin{figure}[h]
    \centering
    \includegraphics[scale=0.35]{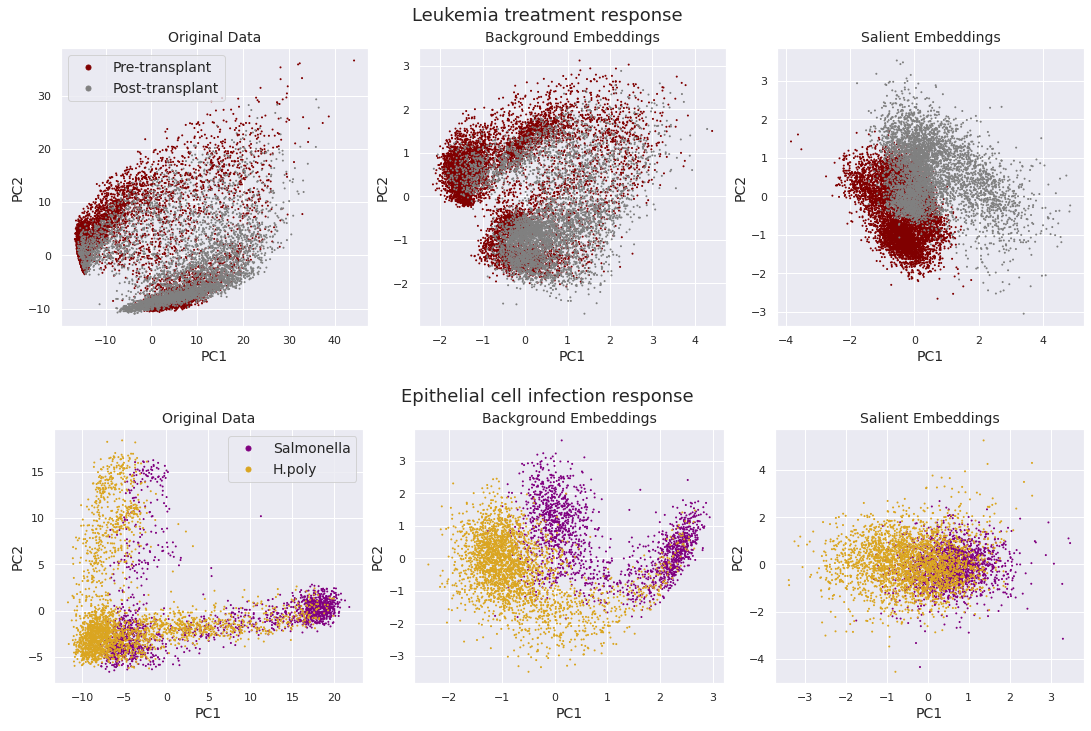}
    \caption{Qualitative RNA seq results for cVAE. Subgroups are not as well separated in the first two PCs of the salient embedding spaces for both datasets as they were with MM-cVAE; this is especially noticeable for the Epithelial cell dataset. Moreover, we see clear separation between subgroups in the first two PCs of the background latent variables for Epithelial cells.}
    \label{fig:qualitative_rna_cvae}
\end{figure}

In Supplementary Figure \ref{fig:qualitative_rna_cvae} we provided qualitative results for cVAE on our two RNA-seq datasets generated using the same procedure as in the main text for MM-cVAE. Here we can see that subgroups are less well-separated in the space spanned by the first two principal components of our salient latent variable values for both datasets than they were for MM-cVAE. This effect is especially pronounced for the Epithelial cell data. Moreover, we find a strong (undesirable) separation between subgroups in the space spanned by the first two PCs for background latent variable values for the Epithelial cell dataset. 

\begin{figure}[h]
    \centering
    \includegraphics[scale=0.35]{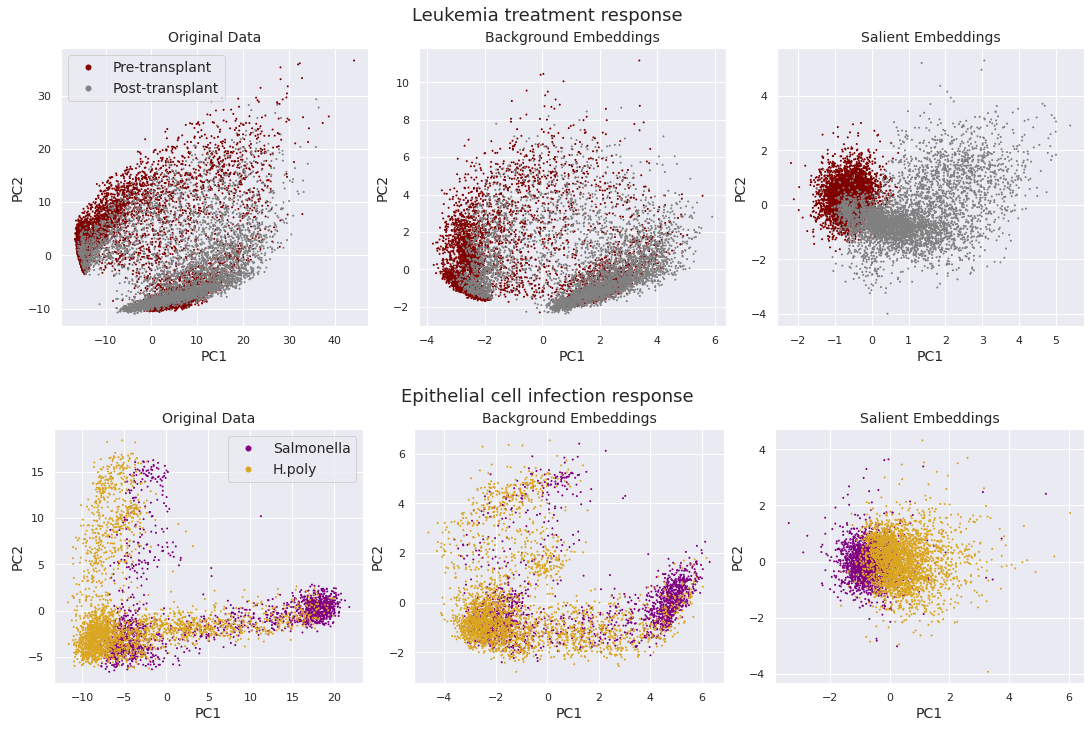}
    \caption{Qualitative RNA seq results for sRB-VAE. While subgroups are (correctly) well mixed in the background latent space, we find more (undesirable) mixing in the salient space than we did with MM-cVAE.}
    \label{fig:qualitative_rna_symmetric}
\end{figure}

In Supplementary Figure \ref{fig:qualitative_rna_symmetric} we provide the corresponding results for sRB-VAE. We similarly find that subgroups are less well separated in the salient latent space for sRB-VAE than they were with MM-cVAE.

\clearpage
\section{Sample Quality Results}

Here we provide the quantitative results measuring sample quality referenced in the main text.

\begin{table}[h]
    \caption{Frechet Inception Distance scores for MM-cVAE and baseline models on the CelebA dataset.}
    \centering
    \input{fid_table}

    \label{table:fid}
\end{table}

\begin{table}[h]
    \caption{Sampled quality as measured by the maximum mean discrepancy for the two RNA-seq datasets for MM-cVAE and baseline models.}
    \centering
    \input{mmd_table}

    \label{table:mmd}
\end{table}

\section{Details on Quantitative Metrics}
 
\subsection{Logistic Regression Score}

For a given set of embeddings, we trained a logistic regression classifier on 80\% of the embeddings, holding out the remaining 20\% for evaluation. In our evaluation, we used scikit-learn's logistic regression solver with the \texttt{max\_iter} parameter set to 1000 and all other parameters left at their default values.  We report the accuracy numbers generated by the \texttt{score} function of our classifier Python objects.

\subsection{Silhouette Score (SS)}

For a given data point $i$, the SS is defined as $SS(i) = \frac{b(i) - a(i)}{\max\{a(i), b(i)\}}$, where $a(i)$ is the average distance from $i$ to other points with the same class label, and $b(i)$ is the average distance to points in the next nearest cluster. For a dataset, the SS is the mean $SS(i)$ for all of the points in that dataset. To calculate the SS, we used the \texttt{silhouette\_score} function in the scikit-learn Python library.

\end{document}

%% file: target_net_new.tex
\begin{tikzpicture}[square/.style={regular polygon,regular polygon sides=4}]

\node[obs] (x) {$\bm{x}_{i}$};

\node[latent, above=of x] (z) {$\bm{z}_{i}$};
\node[latent, right=of z] (s) {$\bm{s}_{i}$};

\edge {z, s} {x};

\plate[inner sep=0.2cm] {pl_tar} {(x)(s)(z)} {$i=1\dots n$} ;
\node[above,font=\large\bfseries] at (current bounding box.north) {Target};

\end{tikzpicture}
\hspace{0.5cm}
\begin{tikzpicture}[square/.style={regular polygon,regular polygon sides=4}]

\node[obs] (x) {$\bm{b}_{j}$};

\node[latent, above=of x] (z) {$\bm{z}_{j}$};
\node[square, draw, right=of z] (s) {$\bm{s'}$};

\edge {z, s} {x};

\plate[inner sep=0.15cm] {pl_tar} {(x)(s)(z)} {$j=1\dots m$} ;
\node[above,font=\large\bfseries] at (current bounding box.north) {Background};

\end{tikzpicture}

%% file: table_background.tex
\begin{tabular}{ c|c|cccc }\toprule
& & \multicolumn{2}{c}{Difference in distribution} & \multicolumn{2}{c}{Difference in distribution}\\
Dataset & Model & \multicolumn{2}{c}{for $\bm{z}_{\bm{x}}$ vs.\
$\bm{z}_{\bm{b}}$} & \multicolumn{2}{c}{for $\bm{s}_{\bm{b}}$ vs.\ $\bm{s}'$}\\
& & \multicolumn{2}{c}{(\textbf{Lower} is better)} & \multicolumn{2}{c}{(\textbf{Lower} is better)}\\
& & \multicolumn{1}{c}{Logistic (\%)} & \multicolumn{1}{c}{Silhouette} & \multicolumn{1}{c}{Logistic (\%)} & \multicolumn{1}{c}{Silhouette}\\
\midrule


& sRB-VAE & $0.72 \pm 0.01$ & $0.02 \pm 0.01$ & $0.83 \pm 0.02$ & $0.35 \pm 0.01$ \\
CelebA  & cVAE & $0.68 \pm 0.01$ & $0.02 \pm 0.00$ & $0.87 \pm 0.06$ & $0.36 \pm 0.01$ \\ 
 & MM-cVAE & $\bm{0.51 \pm 0.01}$ & $\bm{0.00 \pm 0.01}$ & $\bm{0.73 \pm 0.10}$ & $\bm{0.22 \pm 0.05}$ \\
\midrule

 & sRB-VAE & $0.84 \pm 0.01$ & $0.05 \pm 0.01$ & $0.92 \pm 0.05$ & $0.44 \pm 0.06$\\
AML & cVAE & $0.84 \pm 0.03$ & $0.05 \pm 0.01$ & $0.92 \pm 0.05$ & $0.40 \pm 0.04$\\ 
& MM-cVAE & $\bm{0.74 \pm 0.00}$ & $\bm{0.00 \pm 0.01}$ & $\bm{0.53 \pm 0.00}$ & $\bm{0.04 \pm 0.01}$\\
\midrule

& sRB-VAE & $0.75 \pm 0.08$ & $0.01 \pm 0.01$ & $0.95 \pm 0.07$ & $0.47 \pm 0.08$ \\
Epithelial Cells & cVAE & $0.69 \pm 0.04$ & $0.01 \pm 0.01$ & $0.84 \pm 0.04$ & $0.35 \pm 0.01$\\ 
& MM-cVAE & $\bm{0.56 \pm 0.01}$ & $\bm{0.00 \pm 0.01}$ & $\bm{0.51 \pm 0.01}$ & $\bm{0.09 \pm 0.02}$\\
\bottomrule
\end{tabular}

%% file: table_target.tex
\begin{tabular}{ c|c|cccc }\toprule
 &  & \multicolumn{4}{c}{Target Dataset Class Separation} \\
 Dataset & Model &  \multicolumn{2}{c}{$\bm{s}_x$} & \multicolumn{2}{c}{$\bm{z}_{x}$}\\
(Task in target dataset)& & \multicolumn{2}{c}{(\textbf{Higher} is better)} & \multicolumn{2}{c}{(\textbf{Lower} is better)}\\
& & \multicolumn{1}{c}{Logistic (\%)} & \multicolumn{1}{c}{Silhouette} & \multicolumn{1}{c}{Logistic (\%)} & \multicolumn{1}{c}{Silhouette} \\
\midrule

CelebA  & sRB-VAE & $0.79 \pm 0.01$ & $0.11 \pm 0.01$ & $0.83 \pm 0.00$ & $0.05 \pm 0.01$ \\
(Glasses vs.\ caps) & cVAE & $0.75 \pm 0.05$ & $0.06 \pm 0.02$ & $0.81 \pm 0.01$ & $0.04 \pm 0.01$ \\ 
& MM-cVAE & $\bm{0.87 \pm 0.04}$ & $\bm{0.17 \pm 0.03}$ & $\bm{0.58 \pm 0.02}$ & $\bm{0.00 \pm 0.01}$\\
\midrule

AML & sRB-VAE & $0.94 \pm 0.02$ & $0.23 \pm 0.04$ & $0.87 \pm 0.03$ & $0.11 \pm 0.01$\\
(Pre- vs.\ post-transplant & cVAE & $0.88 \pm 0.05$ & $0.18 \pm 0.02$ & $0.77 \pm 0.04$ & $0.06 \pm 0.01$\\ 
patients)& MM-cVAE  & $\bm{0.95} \pm 0.01$ & $\bm{0.28} \pm 0.03$ & $\bm{0.76} \pm 0.02$ & $\bm{0.04} \pm 0.01$\\
\midrule

Epithelial Cells & sRB-VAE & $0.84 \pm 0.03$ & $0.21 \pm 0.05$ & $0.88 \pm 0.03$ & $0.11 \pm 0.01$ \\
(Mice infected with & cVAE & $0.83 \pm 0.10$ & $0.10 \pm 0.05$ & $0.90 \pm 0.04$ & $0.08 \pm 0.02$ \\ 
\textit{Salmonella} vs.\ \textit{H. poly}) & MM-cVAE & $\bm{0.95 \pm 0.03}$ & $\bm{0.33 \pm 0.08}$ & $\bm{0.78 \pm 0.02}$ & $\bm{0.03 \pm 0.01}$\\
\bottomrule
\end{tabular}

%% file: fid_table.tex
    \begin{tabular}{c|l|l}
    \toprule
         \multicolumn{1}{c}{Model} & \multicolumn{1}{c}{Background} &\multicolumn{1}{c}{Target}\\
         \midrule
         sRB-VAE & $335.18 \pm 14.04$ & $299.70 \pm 13.69$\\
         cVAE & $203.05 \pm 4.33$ & $\bm{223.91 \pm 8.83}$ \\
         MM-cVAE & $197.05 \pm 6.44$ & $224.73\pm 5.46$ \\
         VAE & $\bm{174.39 \pm 15.23}$ & $240.38 \pm 3.55$ \\
    \bottomrule
    \end{tabular}

%% file: mmd_table.tex
\begin{tabular}{ c|c|cc }\toprule
Dataset & Model & \multicolumn{2}{c}{Sample Quality (MMD)} \\
& & \multicolumn{1}{c}{Background} & \multicolumn{1}{c}{Target} \\
\midrule


AML & sRB-VAE & $0.37 \pm 0.02$ & $0.22 \pm 0.02$\\
(Pre- vs.\ post-transplant & cVAE & $0.20 \pm 0.01$ & $0.10 \pm 0.01$\\ 
patients)& MM-cVAE & $\bm{0.14 \pm 0.01}$ & $\bm{0.08 \pm 0.01}$ \\
\midrule

Epithelial Cells  & sRB-VAE & $0.20 \pm 0.02$ & $0.22 \pm 0.01$ \\
(Mice infected with & cVAE & $0.15 \pm 0.01$ & $0.13 \pm 0.01$ \\ 
\textit{Salmonella} vs.\ \textit{H. poly})& MM-cVAE & $\bm{0.11 \pm 0.01}$ & $\bm{0.13 \pm 0.01}$ \\
\bottomrule
\end{tabular}

%% file: main.bbl
\begin{thebibliography}{31}
\providecommand{\natexlab}[1]{#1}
\providecommand{\url}[1]{\texttt{#1}}
\expandafter\ifx\csname urlstyle\endcsname\relax
  \providecommand{\doi}[1]{doi: #1}\else
  \providecommand{\doi}{doi: \begingroup \urlstyle{rm}\Url}\fi

\bibitem[Abid and Zou(2019)]{abid2019contrastive}
A.~Abid and J.~Zou.
\newblock Contrastive variational autoencoder enhances salient features.
\newblock \emph{arXiv preprint arXiv:1902.04601}, 2019.

\bibitem[Abid et~al.(2018)Abid, Zhang, Bagaria, and Zou]{abid2018exploring}
A.~Abid, M.~J. Zhang, V.~K. Bagaria, and J.~Zou.
\newblock Exploring patterns enriched in a dataset with contrastive principal
  component analysis.
\newblock \emph{Nature communications}, 9\penalty0 (1):\penalty0 1--7, 2018.

\bibitem[Bengio et~al.(2013)Bengio, Courville, and
  Vincent]{bengio2013representation}
Y.~Bengio, A.~Courville, and P.~Vincent.
\newblock Representation learning: A review and new perspectives.
\newblock \emph{IEEE transactions on pattern analysis and machine
  intelligence}, 35\penalty0 (8):\penalty0 1798--1828, 2013.

\bibitem[Burgess et~al.(2018)Burgess, Higgins, Pal, Matthey, Watters,
  Desjardins, and Lerchner]{burgess2018understanding}
C.~P. Burgess, I.~Higgins, A.~Pal, L.~Matthey, N.~Watters, G.~Desjardins, and
  A.~Lerchner.
\newblock Understanding disentangling in $\beta $-vae.
\newblock \emph{arXiv preprint arXiv:1804.03599}, 2018.

\bibitem[Chen et~al.(2018)Chen, Li, Grosse, and Duvenaud]{chen2018isolating}
T.~Q. Chen, X.~Li, R.~B. Grosse, and D.~K. Duvenaud.
\newblock Isolating sources of disentanglement in variational autoencoders.
\newblock In \emph{NeurIPS}, 2018.

\bibitem[Dziugaite et~al.(2015)Dziugaite, Roy, and
  Ghahramani]{dziugaite2015training}
G.~K. Dziugaite, D.~M. Roy, and Z.~Ghahramani.
\newblock Training generative neural networks via maximum mean discrepancy
  optimization.
\newblock In \emph{Proceedings of the Thirty-First Conference on Uncertainty in
  Artificial Intelligence}, pages 258--267, 2015.

\bibitem[Gretton et~al.(2006)Gretton, Borgwardt, Rasch, Sch{\"o}lkopf, and
  Smola]{gretton2006kernel}
A.~Gretton, K.~Borgwardt, M.~Rasch, B.~Sch{\"o}lkopf, and A.~Smola.
\newblock A kernel method for the two-sample-problem.
\newblock \emph{Advances in neural information processing systems},
  19:\penalty0 513--520, 2006.

\bibitem[Gretton et~al.(2012)Gretton, Borgwardt, Rasch, Sch{\"o}lkopf, and
  Smola]{gretton2012kernel}
A.~Gretton, K.~M. Borgwardt, M.~J. Rasch, B.~Sch{\"o}lkopf, and A.~Smola.
\newblock A kernel two-sample test.
\newblock \emph{The Journal of Machine Learning Research}, 13\penalty0
  (1):\penalty0 723--773, 2012.

\bibitem[Haber et~al.(2017)Haber, Biton, Rogel, Herbst, Shekhar, Smillie,
  Burgin, Delorey, Howitt, Katz, et~al.]{haber2017single}
A.~L. Haber, M.~Biton, N.~Rogel, R.~H. Herbst, K.~Shekhar, C.~Smillie,
  G.~Burgin, T.~M. Delorey, M.~R. Howitt, Y.~Katz, et~al.
\newblock A single-cell survey of the small intestinal epithelium.
\newblock \emph{Nature}, 551\penalty0 (7680):\penalty0 333--339, 2017.

\bibitem[Heusel et~al.(2017)Heusel, Ramsauer, Unterthiner, Nessler, and
  Hochreiter]{heusel2017gans}
M.~Heusel, H.~Ramsauer, T.~Unterthiner, B.~Nessler, and S.~Hochreiter.
\newblock Gans trained by a two time-scale update rule converge to a local nash
  equilibrium.
\newblock \emph{Advances in neural information processing systems}, 30, 2017.

\bibitem[Higgins et~al.(2017)Higgins, Matthey, Pal, Burgess, Glorot, Botvinick,
  Mohamed, and Lerchner]{beta-vae}
I.~Higgins, L.~Matthey, A.~Pal, C.~Burgess, X.~Glorot, M.~Botvinick,
  S.~Mohamed, and A.~Lerchner.
\newblock beta-vae: Learning basic visual concepts with a constrained
  variational framework.
\newblock In \emph{5th International Conference on Learning Representations,
  {ICLR} 2017, Toulon, France, April 24-26, 2017, Conference Track
  Proceedings}. OpenReview.net, 2017.
\newblock URL \url{https://openreview.net/forum?id=Sy2fzU9gl}.

\bibitem[Jones et~al.(2021)Jones, Townes, Li, and
  Engelhardt]{jones2021contrastive}
A.~Jones, F.~W. Townes, D.~Li, and B.~E. Engelhardt.
\newblock Contrastive latent variable modeling with application to case-control
  sequencing experiments.
\newblock \emph{arXiv preprint arXiv:2102.06731}, 2021.

\bibitem[Kim and Mnih(2018)]{kim2018disentangling}
H.~Kim and A.~Mnih.
\newblock Disentangling by factorising.
\newblock In \emph{International Conference on Machine Learning}, pages
  2649--2658. PMLR, 2018.

\bibitem[Kingma and Ba(2014)]{kingma2014adam}
D.~P. Kingma and J.~Ba.
\newblock Adam: A method for stochastic optimization.
\newblock \emph{arXiv preprint arXiv:1412.6980}, 2014.

\bibitem[Kingma and Welling(2014)]{kingma2014autoencoding}
D.~P. Kingma and M.~Welling.
\newblock Auto-encoding variational bayes.
\newblock In Y.~Bengio and Y.~LeCun, editors, \emph{2nd International
  Conference on Learning Representations, {ICLR} 2014, Banff, AB, Canada, April
  14-16, 2014, Conference Track Proceedings}, 2014.
\newblock URL \url{http://arxiv.org/abs/1312.6114}.

\bibitem[Kingma et~al.(2014)Kingma, Mohamed, Rezende, and
  Welling]{kingma2014semi}
D.~P. Kingma, S.~Mohamed, D.~J. Rezende, and M.~Welling.
\newblock Semi-supervised learning with deep generative models.
\newblock In \emph{Advances in neural information processing systems}, pages
  3581--3589, 2014.

\bibitem[Kumar et~al.(2018)Kumar, Sattigeri, and
  Balakrishnan]{kumar2018variational}
A.~Kumar, P.~Sattigeri, and A.~Balakrishnan.
\newblock Variational inference of disentangled latent concepts from unlabeled
  observations.
\newblock In \emph{International Conference on Learning Representations}, 2018.

\bibitem[Li et~al.(2020)Li, Jones, and Engelhardt]{li2020probabilistic}
D.~Li, A.~Jones, and B.~Engelhardt.
\newblock Probabilistic contrastive principal component analysis.
\newblock \emph{arXiv preprint arXiv:2012.07977}, 2020.

\bibitem[Li et~al.(2015)Li, Swersky, and Zemel]{li2015generative}
Y.~Li, K.~Swersky, and R.~Zemel.
\newblock Generative moment matching networks.
\newblock In \emph{International Conference on Machine Learning}, pages
  1718--1727. PMLR, 2015.

\bibitem[Liu et~al.(2015)Liu, Luo, Wang, and Tang]{liu2015deep}
Z.~Liu, P.~Luo, X.~Wang, and X.~Tang.
\newblock Deep learning face attributes in the wild.
\newblock In \emph{Proceedings of the IEEE international conference on computer
  vision}, pages 3730--3738, 2015.

\bibitem[Lopez et~al.(2018)Lopez, Regier, Jordan, and
  Yosef]{lopez2018information}
R.~Lopez, J.~Regier, M.~I. Jordan, and N.~Yosef.
\newblock Information constraints on auto-encoding variational bayes.
\newblock \emph{Advances in Neural Information Processing Systems}, 31, 2018.

\bibitem[Louizos et~al.(2016)Louizos, Swersky, Li, Welling, and
  Zemel]{louizos2016fair}
C.~Louizos, K.~Swersky, Y.~Li, M.~Welling, and R.~S. Zemel.
\newblock The variational fair autoencoder.
\newblock In Y.~Bengio and Y.~LeCun, editors, \emph{4th International
  Conference on Learning Representations, {ICLR} 2016, San Juan, Puerto Rico,
  May 2-4, 2016, Conference Track Proceedings}, 2016.
\newblock URL \url{http://arxiv.org/abs/1511.00830}.

\bibitem[Mathieu et~al.(2019)Mathieu, Rainforth, Siddharth, and
  Teh]{mathieu2019disentangling}
E.~Mathieu, T.~Rainforth, N.~Siddharth, and Y.~W. Teh.
\newblock Disentangling disentanglement in variational autoencoders.
\newblock In \emph{International Conference on Machine Learning}, pages
  4402--4412. PMLR, 2019.

\bibitem[Paige et~al.(2017)Paige, van~de Meent, Desmaison, Goodman, Kohli,
  Wood, Torr, et~al.]{paige2017learning}
B.~Paige, J.-W. van~de Meent, A.~Desmaison, N.~Goodman, P.~Kohli, F.~Wood,
  P.~Torr, et~al.
\newblock Learning disentangled representations with semi-supervised deep
  generative models.
\newblock \emph{Advances in Neural Information Processing Systems},
  30:\penalty0 5925--5935, 2017.

\bibitem[Ruiz et~al.(2019)Ruiz, Martinez, Binefa, and
  Verbeek]{ruiz2019learning}
A.~Ruiz, O.~Martinez, X.~Binefa, and J.~Verbeek.
\newblock Learning disentangled representations with reference-based
  variational autoencoders.
\newblock \emph{arXiv preprint arXiv:1901.08534}, 2019.

\bibitem[Severson et~al.(2019)Severson, Ghosh, and
  Ng]{severson2019unsupervised}
K.~A. Severson, S.~Ghosh, and K.~Ng.
\newblock Unsupervised learning with contrastive latent variable models.
\newblock In \emph{Proceedings of the AAAI Conference on Artificial
  Intelligence}, volume~33, pages 4862--4869, 2019.

\bibitem[Shaham et~al.(2017)Shaham, Stanton, Zhao, Li, Raddassi, Montgomery,
  and Kluger]{shaham2017removal}
U.~Shaham, K.~P. Stanton, J.~Zhao, H.~Li, K.~Raddassi, R.~Montgomery, and
  Y.~Kluger.
\newblock Removal of batch effects using distribution-matching residual
  networks.
\newblock \emph{Bioinformatics}, 33\penalty0 (16):\penalty0 2539--2546, 2017.

\bibitem[van~den Oord et~al.(2017)van~den Oord, Vinyals, and
  Kavukcuoglu]{van2017neural}
A.~van~den Oord, O.~Vinyals, and K.~Kavukcuoglu.
\newblock Neural discrete representation learning.
\newblock In \emph{Proceedings of the 31st International Conference on Neural
  Information Processing Systems}, pages 6309--6318, 2017.

\bibitem[Zhao et~al.(2019)Zhao, Song, and Ermon]{zhao2019infovae}
S.~Zhao, J.~Song, and S.~Ermon.
\newblock Infovae: Balancing learning and inference in variational
  autoencoders.
\newblock In \emph{Proceedings of the aaai conference on artificial
  intelligence}, volume~33, pages 5885--5892, 2019.

\bibitem[Zheng et~al.(2017)Zheng, Terry, Belgrader, Ryvkin, Bent, Wilson,
  Ziraldo, Wheeler, McDermott, Zhu, et~al.]{zheng2017massively}
G.~X. Zheng, J.~M. Terry, P.~Belgrader, P.~Ryvkin, Z.~W. Bent, R.~Wilson, S.~B.
  Ziraldo, T.~D. Wheeler, G.~P. McDermott, J.~Zhu, et~al.
\newblock Massively parallel digital transcriptional profiling of single cells.
\newblock \emph{Nature communications}, 8\penalty0 (1):\penalty0 1--12, 2017.

\bibitem[Zou et~al.(2013)Zou, Hsu, Parkes, and Adams]{zou2013contrastive}
J.~Y. Zou, D.~Hsu, D.~C. Parkes, and R.~P. Adams.
\newblock Contrastive learning using spectral methods.
\newblock \emph{Proceedings of Advances in Neural Information Processing
  Systems}, 2013.

\end{thebibliography}
